\documentclass[sigconf]{acmart}
\usepackage{xcolor}
\usepackage{float}
\usepackage{caption}
\usepackage{subcaption}
\usepackage{url}
\usepackage{enumitem}
\usepackage{multirow}
\usepackage{verbatim} 

\begin{document}

\title{Adapting to Skew: Imputing Spatiotemporal Urban Data \\ with 3D Partial Convolutions and Biased Masking}


\author{Bin Han}
\email{bh193@uw.edu}
\affiliation{%
  \institution{University of Washington}
  \city{Seattle}
  \country{USA}
}

\author{Bill Howe}
\email{billhowe@uw.edu}
\affiliation{%
  \institution{University of Washington}
  \city{Seattle}
  \country{USA}
}

\begin{abstract}
We adapt image inpainting techniques to impute large, irregular missing regions in urban settings characterized by sparsity, variance in both space and time, and anomalous events. Missing regions in urban data can be caused by sensor or software failures, data quality issues, interference from weather events, incomplete data collection, or varying data use regulations; any missing data can render the entire dataset unusable for downstream applications. To ensure coverage and utility, we adapt computer vision techniques for image inpainting to operate on 3D histograms (2D space + 1D time) commonly used for data exchange in urban settings.

Adapting these techniques to the spatiotemporal setting requires handling skew: urban data tend to follow population density patterns (small dense regions surrounded by large sparse areas); these patterns can dominate the learning process and fool the model into ignoring local or transient effects.  To combat skew, we 1) train simultaneously in space and time, and 2) focus attention on dense regions by biasing the masks used for training to the skew in the data. We evaluate the core model and these two extensions using the NYC taxi data and the NYC bikeshare data, simulating different conditions for missing data. We show that the core model is effective qualitatively and quantitatively, and that biased masking during training reduces error in a variety of scenarios. We also articulate a tradeoff in varying the number of timesteps per training sample: too few timesteps and the model ignores transient events; too many timesteps and the model is slow to train with limited performance gain.
\end{abstract}

\ccsdesc[500]{General and reference~Empirical studies}
\ccsdesc[500]{Computing methodologies~Computer vision}
\ccsdesc[500]{Applied computing}

\keywords{image inpainting, urban computing, spatial-temporal, missing data}

\maketitle

\section{Introduction}\label{introduction}

High-quality, longitudinal, and freely available urban data, coupled with advances in machine learning, improve our understanding and management of urban environments. Although conventional machine learning techniques are common in urban applications~\cite{vogel2011understanding, yoon2012cityride, moreira2013predicting}, neural architectures are opening new opportunities by adapting convolutional, recurrent, and transformer architectures to spatiotemporal settings~\cite{bill-bin-2022, liao2018deep, ma2017learning, shen2018stepdeep, yuan2018hetero, yao2018modeling, zhang2018predicting, zhang2016dnn}; see Grekousis 2020 for a recent survey \cite{grekousis2019artificial}.  For example, spatio-temporal neural architectures have been used in predictions of rideshare demand \cite{rideshare1, rideshare2}, traffic conditions \cite{traffic1, traffic2}, and air quality~\cite{aircondition1, aircondition2}.  But these models depend on access to complete, longitudinal datasets.  Such datasets are inconsistent in availability and quality, limiting the opportunity for understanding cities as the complex systems they are~\cite{allen2012cities,ha2021unraveling,kempeswest2020,west2017scalebook}.
\begin{figure}[!ht]
    \centering
        \includegraphics[width=0.48\textwidth]{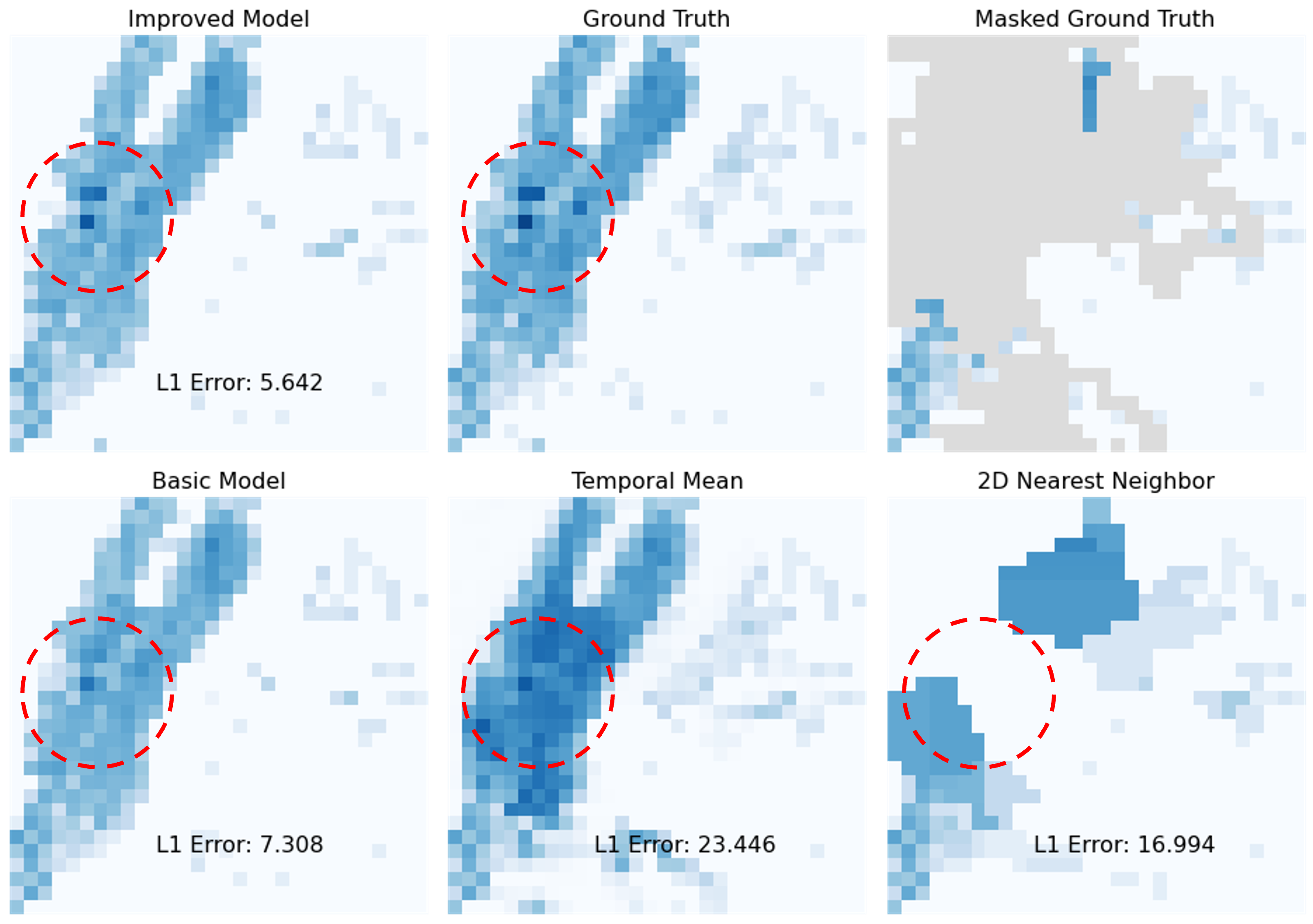}
        \caption{A histogram of taxi pickups in Manhattan.  We adapt imagine inpainting techniques to reconstruct missing and corrupted data in urban settings: The improved model (upper left) uses biased masking and temporal context to capture local effects (red circle).  The basic model (lower left) uses ordinary masking and is insensitive to local effects.  Baseline methods that ignore space (lower middle) or time (lower right) are not competitive.  Classical linear methods such as kriging and inverse-distance weighting (not shown) cannot impute large irregular regions in dynamic settings.  
        }
    \label{fig:teaser}
\end{figure}

This inconsistency persists despite significant investments in open data.  Over the last two decades, cities have increasingly released datasets publicly on the web, proactively, in response to transparency regulation. For example, in the US, all 50 states and the District of Columbia have passed some version of the federal Freedom of Information (FOI) Act. While this first wave of open data was driven by FOI laws and made national government data available primarily to journalists, lawyers, and activists, a second wave of open data, enabled by the advent of open source and web 2.0 technologies, was characterized by an attempt to make data ``open by default" to civic technologists, government agencies, and corporations~\cite{verhulst2020emergence}. While  open data has indeed made significant data assets available online, their uptake and use has been weaker than anticipated~\cite{verhulst2020emergence}, an effect attributable to convenience sampling effects~\cite{lee2015open}: We release what we can, even if portions are missing, corrupt, or anomalous.

In this paper, we consider a neural data cleaning strategy based on masking out corrupted regions and using a trained model to reconstruct the masked region.  These masks are necessarily large, irregular, and extend in both time and space; they may represent political boundaries (municipal zoning, zip codes, city blocks), sensor or software failures \cite{ZHANG2019337,LI201464,zhang-missingdata}, varying legal restrictions \cite{homicide, cybersecurity}, or unusual events (adverse weather). These missing patches can destroy the utility of the entire dataset for applications that assume coverage.  By modeling missing or corrupted data by an arbitrary mask, we afford user control: any areas can be masked and reconstructed, regardless of the reason. We envision tools to improve the coverage and quality of data for use in downstream urban learning tasks~\cite{rideshare1,rideshare2,traffic1,traffic2,aircondition1, aircondition2}.

Following the literature, we represent spatiotemporal event data in a 2D or 3D raster form (e.g., a histogram). Our basic model uses the partial convolution approach from Liu et al~\cite{liu2018irregular} to handle the irregular boundaries of missing data (e.g., districts), which focuses model attention on the valid regions while shrinking the masked region, layer-by-layer, to obtain a complete prediction. More recent approaches to image inpainting on the web emphasize eliminating perceptual artifacts rather than numerical accuracy and are therefore less relevant to our setting.  Our contribution is to extend the basic model to the 3D spatiotemporal setting and propose a training regime that adapts to the skewed distribution found in practice.

Spatiotemporal interpolation of missing data has been widely studied in the earth sciences~\cite{sluiter09,oceaninterp02}, especially in remote sensing where weather effects can obscure measurement~\cite{stock20,zhang-missingdata}.  Conventional statistical approaches to impute missing values, such as global/local mean imputation, interpolation, and kriging, are essentially linear, and therefore limited in their ability to capture the nonlinear dynamics needed to impute large irregular missing regions. Neural image inpainting techniques~\cite{liu2018irregular,yu2019freeform} can recover missing patches via training on large datasets of independent images, such that the reconstructed images appear realistic. These approaches have shown promising results with global climate data~\cite{tasnim-missingdata}, but have not been adapted to the urban setting in which data are not smooth functions of space and time, but are rather histograms of events constrained by the built environment.

\begin{figure}[!ht]
    \centering
        \includegraphics[width=0.48\textwidth]{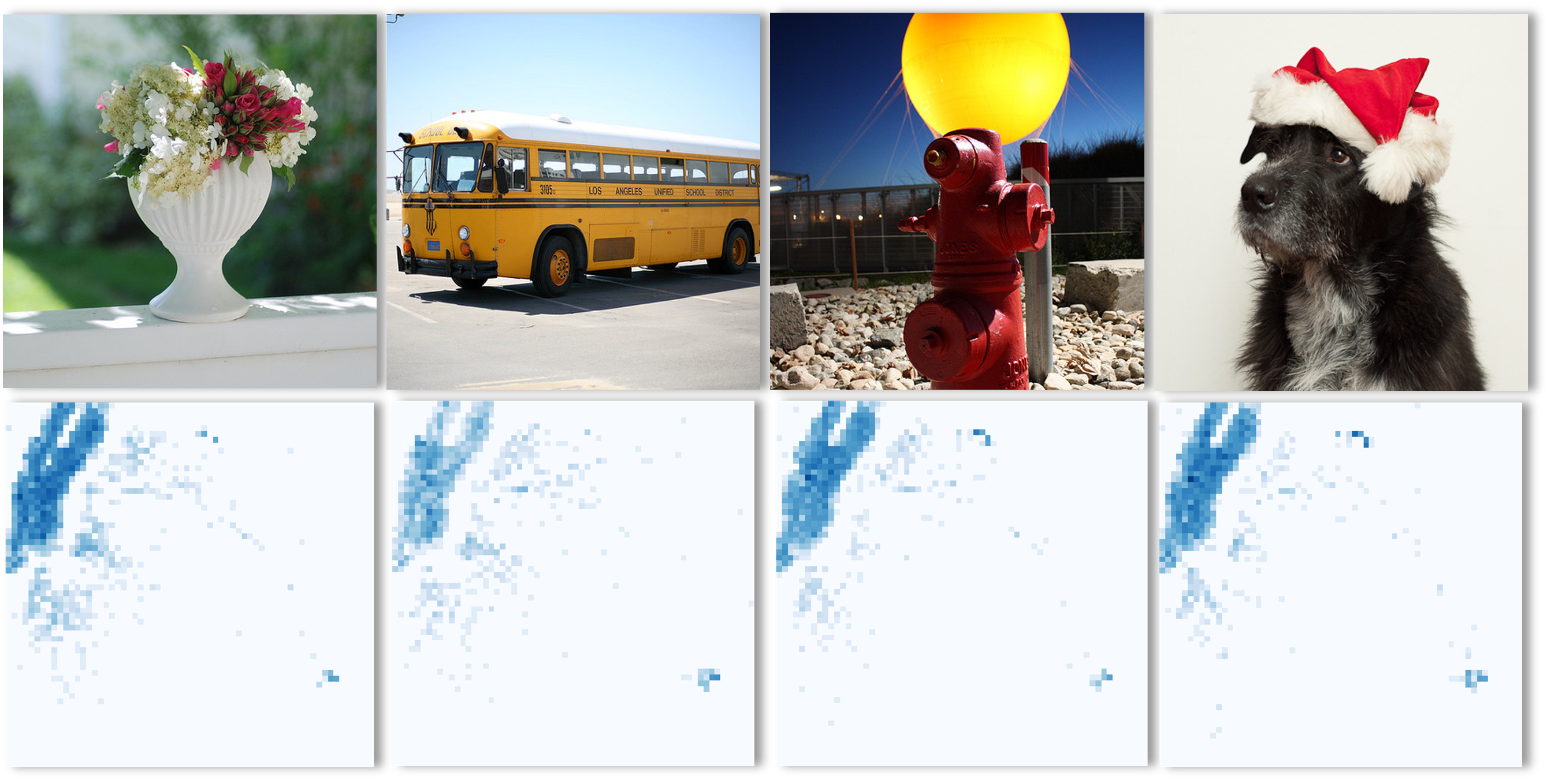}
        \caption{Urban data (bottom row) exhibits skewed, sparse, yet stable distributions that can dominate learning, in contrast with the diversity of natural images (top row).}
    \label{image-comparison}
    \vspace{-1em}
\end{figure}

The goal of inpainting for natural images is to produce a subjectively recognizable image free from perceptible artifacts.  But the goal in our setting is quantitative accuracy: we intend for our reconstructed results to be used numerically in downstream applications.  The distribution is relatively stable, but exhibits skew and sparsity that can obscure local, dynamic features (Figure \ref{image-comparison}). 

The challenge for imputation in the urban setting is \emph{skew}: urban data tend to follow population density patterns --- small dense regions surrounded by large sparse areas.   These population patterns can dominate the learning process and fool the model into ignoring numerical accuracy in dense regions, even while aggregate error may remains low. To combat skew, we 1) bias the training process to focus on populated regions by seeding the mask in non-zero areas; (2) use 3D convolutions and vary the number of timesteps in each 3D training sample to capture transient events.  Together, these two techniques complement each other: biased masking focuses attention on dense regions, and 3D convolutions with a large chunk size focus attention on sparse regions.

We evaluate these techniques on the NYC taxi data (a popular dataset for its coverage and quality) and a NYC bikeshare dataset (less dominated by the built environment). We find that the basic model is effective for urban data imputation, while biased masking reliably reduces error over random masking, both globally and locally. Additionally, we find that the number of timesteps per training sample exhibits a tradeoff: too few timesteps and the model ignores transient patterns, while too many timesteps significantly increases training time without enhancing the inpainting results. We evaluate specific local scenarios (high-traffic locations, low-traffic locations, high-variability locations, anomalous events) to reflect the use cases distinct from image inpainting on the web (where subjective quality is all that matters).

In summary, we make the following contributions:
\begin{itemize}[leftmargin=0.12in]
    \item We evaluate a basic model adapting image inpainting techniques to urban histograms characterized by skew and sparsity effects due to constraints by the built environment, demonstrating qualitative and quantitative accuracy relative to classical methods.
    
    \item We improve on this basic model by extending to the 3D spatiotemporal setting to better recognize transient events; we analyze the training time and performance tradeoffs of varying the number of timesteps per training sample.
    
    \item We propose a self-supervised training process called biased masking to encourage the model to attend to dense population regions and thereby improve accuracy on the highly dynamic regions typical in urban environments; we show that biased masking reliably improves convergence.
    
    \item We evaluate these techniques on two real mobility datasets (NYC taxi trips and NYC bikeshare trips), both globally and locally in varying traffic conditions, weather events, and disruptions.  Finally, we show that the model can be used to remove or synthesize anomalous events through targeted masking.
\end{itemize}

\section{Related Work}\label{literature}
Our work is informed by techniques in image inpainting and geospatial interpolation.

\textbf{Image Inpainting}\label{inpainting}
Image inpainting, or image completion, is a task of synthesizing missing pixels in images, such that the reconstructed images are visually credible and semantically realistic. In computer vision, there are two broad categories of inpainting techniques. The first category contains diffusion-based or patch-based methods, which utilize low-level image features to recover the missing pixels. The second category contains learning-based methods that generally involve the training of deep neural networks. 

Diffusion-based methods \cite{Bertalmo2000ImageI, Ballester2001, levin_global_statistics} propagate information from neighboring valid pixels to missing pixels, typically from border to the center of the missing regions. Those techniques are convenient to apply, but are limited to small missing regions. Recently, Saharia et al. \cite{saharia2022palette} developed an image-to-image translation framework based on conditional diffusion models. The evaluation on inpainting task outperformed several learning-based methods. Patch-based inpainting techniques \cite{BerTalmio_structureImageI, Drori2003, James2007,Darabi2012} function by searching similar patches from the valid regions of the same image or from other images, and then paste the patches to the target missing region. However, this process could induce high computational costs. A milestone of patch-based approach, PatchMatch \cite{Barnes2009}, speeds up the search process with a new nearest neighbor algorithm. 

Learning-based methods are trained to learn image patterns with large volume of image data, thus being capable of recovering missing regions, as well preserving the semantics of the imagery. Pathak et al.\cite{pathak2016context} proposed context encoder, which was the first work to combine CNN with generative adversarial network. It applied the encoder-decoder architecture and used both $\ell_2$ reconstruction loss and generative adversarial loss in the objective function. Lizuka et al. \cite{globalyandlocally} improved on their work by incorporating global and local discriminator, which improved content consistency between the valid and missing region. Additionally, they replaced general convolutional layers with dialated convolutional layers to better capture information from distant pixels. Yu et al. \cite{yu2018generative} proposed proposed a two-stage coarse-to-fine model architecture and incorporated contextual attention layer to attend to related features from spatially distant regions. They also replaced general generative adversarial loss with WGANS loss. Liu et al. \cite{liu2018irregular} proposed partial convolution, allowing inpainting models to be used on irregular holes rather than just rectangular missing regions. On top the work of partial convolution, Yu et al. \cite{yu2019freeform} proposed gated convolutional layers to automatically learn and update the masks as opposed to rule-based update. To further address the problems of blurry textures and distorted structures in the inpainted images, Liu et al. \cite{liu2019csa} proposed coherent semantic attention layer, which can both preserve contextual structure and capture semantic relevance between hole features. Zhou et al.\cite{zhou2020learning} incorporated dual spatial attention modules into the U-Net architecture, which can capture the correlations between facial textures at different scales. Seven different discriminators are utilized to ensure realistic local details as well as global consistency. Yu et al. \cite{yu2020region} designed spatial region-wise normalization (RN) to overcome the problem of mean and variance shifts. RN computes the mean and variances separately for the missing and valid regions. Xu et al. \cite{xu2021texture} combined the paradigms of both patch-based and learning-based methods, and inpainted missing regions using textures of patch samples from unmasked regions. Additionally, they proposed patch distribution loss to ensure the quality of synthesized missing regions. Zeng et al. \cite{zeng2022aggregated} introduced aggregated contextual transformation GAN, aiming to improve content reasoning from distant pixels and enhance details of synthesized textures. For more image inpainting works, we refer reader to the following surveys \cite{qin2021image, jam2021comprehensive, liu2022overview}.

The recent trajectory in image inpainting involves reducing or eliminating perceptual artifacts such as discontinuous edges and blurred patches using new loss terms, image preprocessing, or training regimes that favor subjective quality over numerical accuracy. For example, the work of Liu et al.\cite{liu2019csa}, Yu et al. \cite{yu2020region}, and Xu et al. \cite{xu2021texture} all propose extensions to partial convolutions to repair blurred boundaries between missing and valid regions. Since our focus is on numerical accuracy and downstream utility of the synthesized data, we base our approach on partial convolutions from Liu et al. \cite{liu2018irregular}.  Additionally, we aim to design and study architecture-agnostic training regimes that can be used with newer models when applicable. 

\textbf{Geospatial Missing Data Imputation}\label{inpaint-app}
Classical spatio-temporal interpolation methods, generally variants of inverse-distance or nearest-neighbor weighting~\cite{knn-icu, knn-pm10}, kriging~\cite{cokrigging, krigging-sequence}, or matrix factorization~\cite{gong2021spatial} are variations of linear methods that do not attempt to (and cannot) interpolate within large, arbitrary, irregular regions, and typically do not seamlessly consider both space and time.
Physics-based models based on computational fluid dynamics~\cite{BLOCKEN2015219} or agent-based models that directly encode human behavior \cite{individual, SOUZA2019858} have been used to infer mobility dynamics, but must be designed separately for each application rather than learned automatically from data. 
Gong et al.~\cite{gong2021spatial} solve multi-variable non-negative matrix factorization to impute urban data, but assume the availability of multiple variables and do not consider arbitrary irregularities. Zhang et al. \cite{zhang-missingdata} were concerned about the malfunction of satellites and poor atmospheric conditions (e.g. thick cloud), which could produce missing regions in remote sensing data. They proposed unified spatial-temporal-spectral deep CNN architecture to recover the missing information in satellite images. Kang et al. \cite{kang-missingdata} modified the architecture from \cite{yu2018generative} to restore the missing patterns of sea surface temperature (SST) from satellite images. Tasnim and Mondal \cite{tasnim-missingdata} also adopted the coarse-to-fine inpainting architecture from \cite{yu2018generative} to restore satellite images. The innovation of their work is the abandonment of coarse-inpainting pipeline. Instead, they used another highly correlated temporal image as an auxiliary input to go through the refinement pipeline. Additionally, Kadow, Hall and Ulbrich \cite{kadow-missingdata} borrowed the architecture from \cite{liu2018irregular} to reconstruct missing climate information. 
In the geo-spatial domain, most of the literature that we found applied image inpainting techniques on remote sensing data. As far as we acknowledge, there is no prior work that has taken advantage of image inpainting methods to reconstruct missing values in urban data. 

\section{Representative Datasets}\label{data}
We worked with two mobility datasets: NYC taxi data and NYC bikeshare data. Although potential \emph{applications} of the proposed model are widely available, datasets on which to \emph{evaluate} the model are rare: we need longitudinal coverage to provide ground truth, sufficient complexity to study both global and local fidelity, and accessibility to a general audience for expository purposes.  Mobility data achieves all three goals. 
\begin{itemize}[leftmargin=0.12in]
    \item \textbf{NYC Taxi Data}. NYC taxi trip data were collected from NYC Open Data portal from 2011 to 2016\footnote{\url{https://opendata.cityofnewyork.us/data/}}. The year 2011 --- 2015 cover the trips throughout the entire year, while 2016 only covers the first half of the year until June 30. The raw data are presented in tabular format. Each record from the data summarizes the information for one single taxi trip, which contains the longitude and latitude of the location where the taxi took off. Each record can be viewed as one taxi demand count.
    
    \item \textbf{NYC Bikeshare Data}: NYC bikeshare data were collected from NYC DOT from 2019 to 2021 portal.\footnote{\url{https://ride.citibikenyc.com/system-data}} All three datasets cover the bike trips throughout the entire year. Similar to the taxi data, the raw data are presented in tabular format. Each data point summarizes the information for one single bike trip, including the longitude and latitude of the location where the bike was unlocked. Each record can be viewed as one bike demand count.
\end{itemize}
\begin{figure}[!ht]
    \centering
        \includegraphics[width=0.44\textwidth]{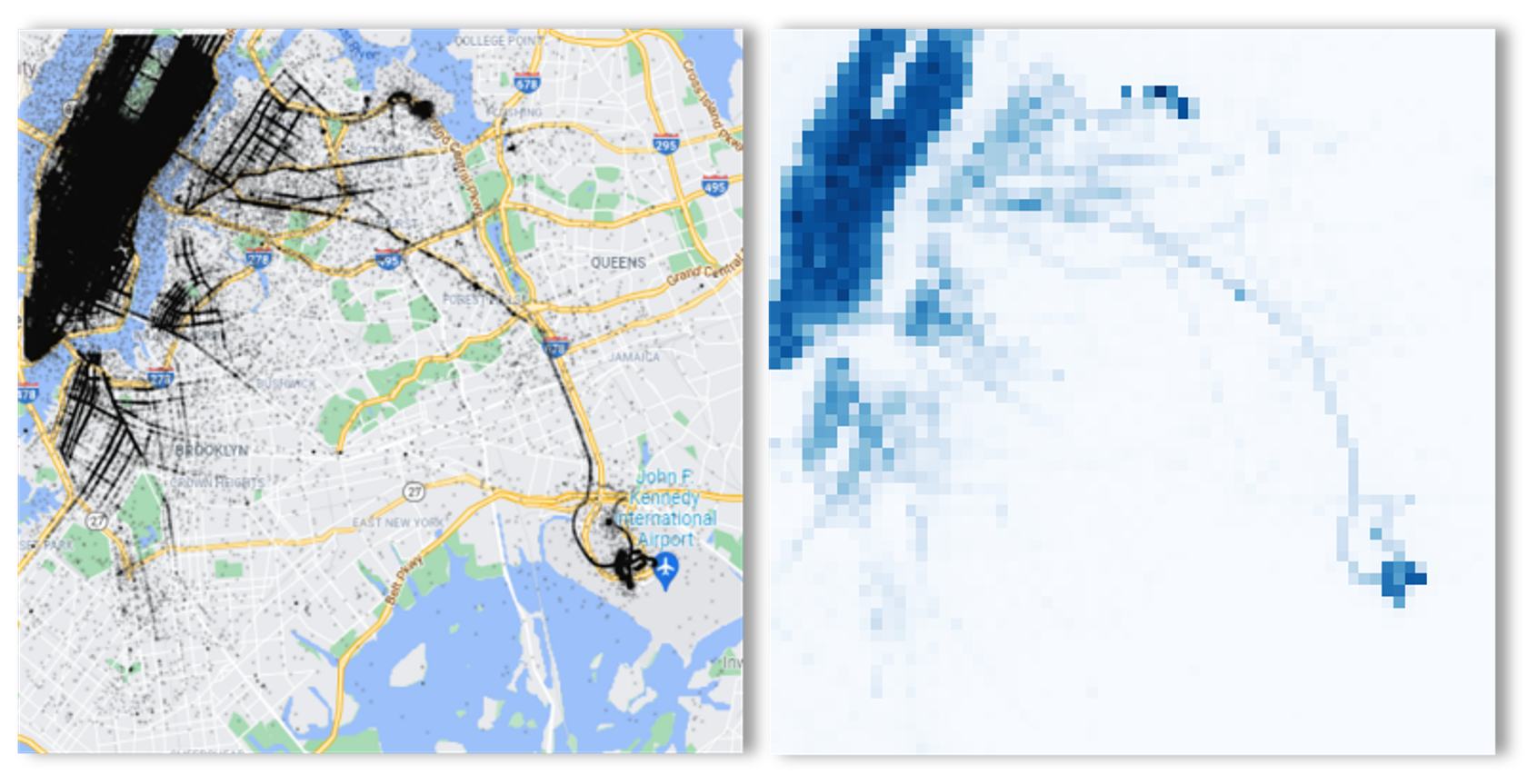}
        \caption{Left: Taxi pickups in 2011 overlaid on a regional map. The distribution of taxi demand count is skewed --- high demand in Manhattan, and low demand in the surrounding areas. Right: Taxi pickups aggregated into a 64$\times$64 histogram. }
    \label{nyc_taxi}
\end{figure}
We aggregate both datasets into a 3D histogram by defining a rectangular region, then binning mobility events into a regular grid to create a 2D histogram amenable to image techniques. These 2D images are stacked to create 3D blocks.  The temporal depth of the block is a parameter we study in this paper. We defined the NYC region as shown in Figure \ref{nyc_taxi}. At this resolution, the region has a relatively balanced coverage of areas with different levels of demand counts: high demand in Manhattan and near airports, and low demand elsewhere. This skew is common in urban applications and represents both an opportunity and a challenge for neural prediction: the patterns are relatively stable, but the sparse regions can dilute the learning process.  

We then define a 64$\times$64 grid over the region of interest. Then, for each hour of each day, we count the number of taxi/bike trips that began within each pixel. The value is commonly interpreted as an estimate of demand.  We do not consider multiple resolutions in this paper. After processing, we have 30,648 training images and 3,620 test images in NYC taxi data, and we have 25,560 training images and and 720 test images in NYC bikeshare data. Figure \ref{nyc_taxi} shows the defined region and an example of the corresponding taxi demand histogram. 

\section{Inpainting Model}\label{problem-setup}
In this section, we describe the basic model for using partial convolutions for inpainting spatiotemporal urban histograms. Each sample consists of a \emph{masked} region with unknown, corrupted, or inaccurate values to be reconstructed and a \emph{valid} region with known values.   The task is to predict values in the masked region to match the original image. Training is self-supervised by creating random masks for any input image; we consider the manner in which the masks are created in this paper. 

\subsection{Model Architecture}\label{architecture}
We adapt the architecture from Liu et al.\cite{liu2018irregular}, which proposed partial convolutional layers to accommodate irregular masks. Partial convolutions ignore the masked region, but the mask is updated after each partial convolution layer: after several partial convolution layers, all the values in the mask will be set to one such that the entire output is considered valid.  We use the U-Net architecture with skip connections \cite{unet}, with all the convolution layers replaced with partial convolution layers. 
\begin{figure}[!hb]
    \centering
        \includegraphics[width=0.44\textwidth]{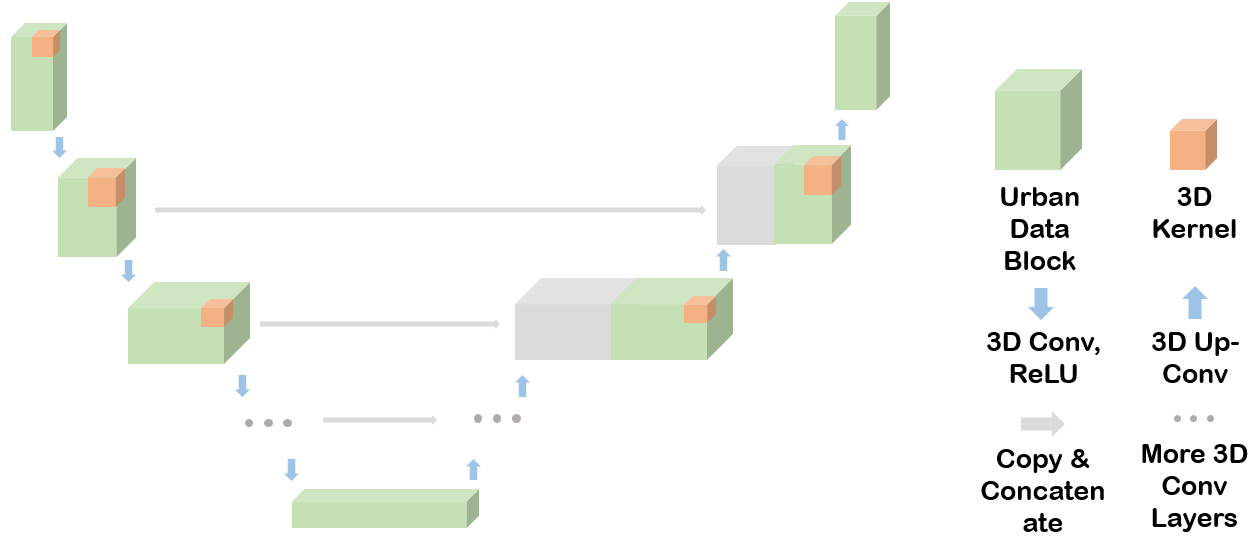}
        \caption{The model architecture is a U-Net extended to 3D, partial convolutional layers~\cite{liu2018irregular} to ignore masked regions during training. In the decoding branch, multiple 3D up-convolutional layers are utilized and skip-connections are applied. In total, there are six encoding layers and six decoding layers.}
    \label{model_architecture}
\end{figure}
Web images only contain 2D information (ignoring RGB channels), but urban histograms vary in both space and time.  As a result, our training data is essentially one massive 3D block rather than a large number of independent training images.  We therefore have a design choice of how to ``shred'' this block into training samples.  In this paper, we consider only the temporal extent in 3D;  varying spatial resolution, bounds, or overlap during rasterization of the source data is left for future work.   

If we slice the input into individual timesteps, the model cannot exploit temporal consistency. We therefore extend all convolutional layers, inputs, and masks, to 3D, and consider the effect of varying the number of timesteps per training sample.  The inputs are 3D image blocks of dimension $T\times W\times H$, where $T$ represents the temporal dimension. The masks are also in 3D blocks with the same shape as the image block. The model architecture is illustrated in Figure \ref{model_architecture}. The parameters of each convolutional layer appear in Table \ref{conv_layers}.
\begin{table}[!hb]
    \small
    \tabcolsep=0.22cm
    \renewcommand*{\arraystretch}{0.9}
    \centering
    \begin{tabular}{|c|c|c|c|c|}
        \hline
        \textbf{Layers} & \textbf{Channel} & \textbf{Kernel Size} & \textbf{Stride} & \textbf{Padding} \\\hline
        encoder 1 & 64 & (1,3,3) & (1,2,2) & (0,1,1) \\ 
        encoder 2 & 128 & (1,3,3) & (1,2,2) & (0,1,1) \\ 
        encoder 3 & 256 & (1,3,3) & (1,2,2) & (0,1,1) \\ 
        encoder 4 & 512 & (1,3,3) & (1,2,2) & (0,1,1) \\ 
        encoder 5 & 512 & (T,3,3) & (2,2,2) & (2*((T-1)//4),1,1) \\ 
        encoder 6 & 512 & (T,3,3) & (2,2,2) & (2*((T-1)//4),1,1) \\\hline
        decoder 1 & 512 & (1,3,3) & (1,1,1) & (0,1,1) \\ 
        decoder 2 & 512 & (1,3,3) & (1,1,1) & (0,1,1) \\ 
        decoder 3 & 256 & (1,3,3) & (1,1,1) & (0,1,1) \\ 
        decoder 4 & 128 & (1,3,3) & (1,1,1) & (0,1,1) \\ 
        decoder 5 & 64 & (1,3,3) & (1,1,1) & (0,1,1) \\ 
        decoder 6 & 1 & (1,3,3) & (1,1,1) & (0,1,1) \\ \hline 
    \end{tabular}
    \caption{Parameters of 3D convolutional layers. T represents the temporal dimension of the image block.}
    \label{conv_layers}
\end{table}
\vspace{-0.5cm}
\subsection{Loss function}\label{loss}
We used $\ell_1$ loss as the objective function  for pixel-wise reconstruction accuracy. The $\ell_1$ loss term bridges the absolute gap between the reconstructed value and the ground truth. 
We adopt the following notation
\begin{itemize}[leftmargin=0.12in,label={}]
    \item $\textbf{I}_{gt} \in \mathbb{R}^{T\times W\times H}$: the block of ground truth images. $T$ represents the temporal dimension of the block. 
    \item $\textbf{I}_{out} \in \mathbb{R}^{T\times W\times H}$: the block of reconstructed images. 
    \item $\textbf{M} \in \mathbb{R}^{T\times W\times H}$: the block of binary masks. 
    \item $N_{\textbf{I}} = T*W*H$: the total number of pixels in the image block.
    \item $N_{\textbf{valid}}$: the total number of valid pixels in the image block.
    \item $N_{\textbf{hole}}$: the total number of missing pixels in the image block.
\end{itemize}

Following Liu, we separate the valid and hole regions in the $\ell_1$ loss. Even though the valid region has available data and we therefore typically would not use the predicted values in practice, we want to include this loss during training to improve continuity across mask boundaries (and therefore improve overall error).  The $\ell_1$ loss is calculated as
\begin{displaymath}
\mathcal{L}_{total} = \mathcal{L}_{valid} + \lambda\mathcal{L}_{hole}
\end{displaymath}
\noindent where
\begin{align*}
\mathcal{L}_{hole} &= \frac{1}{N_\textbf{hole}}||(1-\textbf{M})\odot(\textbf{I}_{out}-\textbf{I}_{gt})||_\textbf{1}\\ \mathcal{L}_{valid} &= \frac{1}{N_{\textbf{valid}}}||\textbf{M}\odot(\textbf{I}_{out}-\textbf{I}_{gt})||_\textbf{1}
\end{align*}

\subsection{Biased Masking}\label{mask}
\begin{figure}[!ht]
    \centering
     \begin{subfigure}[b]{0.46\textwidth}
         \centering
         \includegraphics[width=\textwidth]{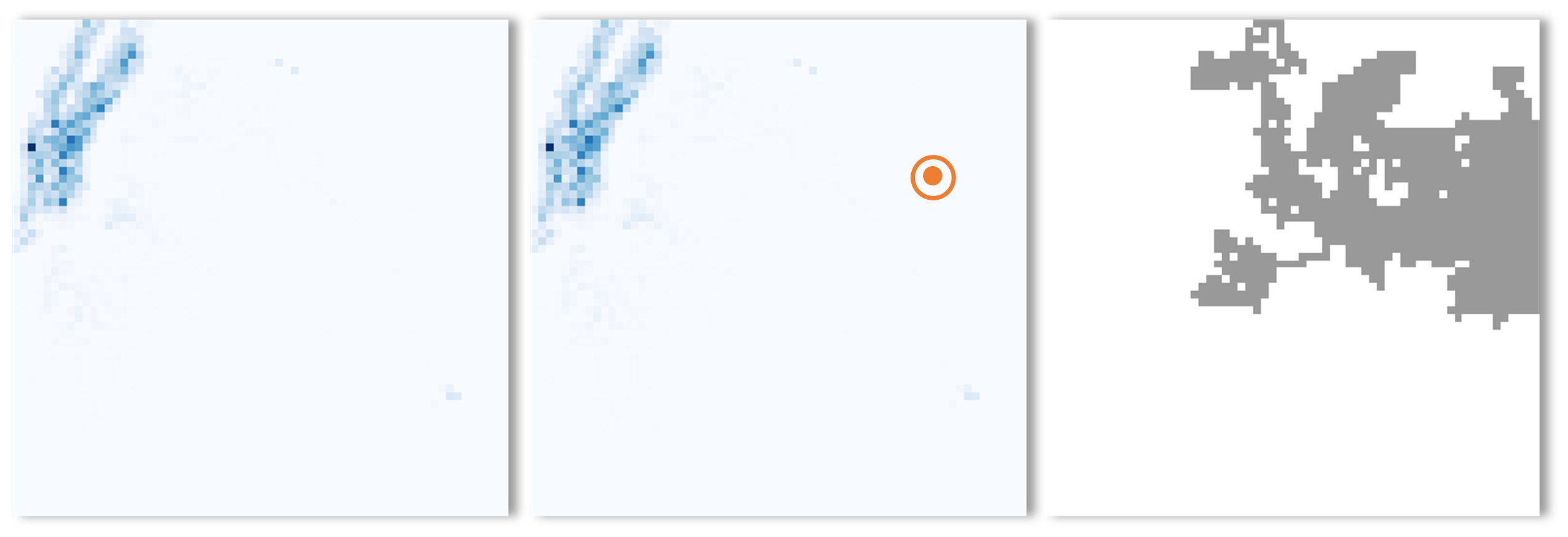}
         \caption{Random masking. For each image (left), randomly select a starting point (orange dot, middle), then grow a mask via random walk to generate a masked region (right).}
         \label{random_mask}
     \end{subfigure}
     \begin{subfigure}[b]{0.46\textwidth}
         \centering
         \includegraphics[width=\textwidth]{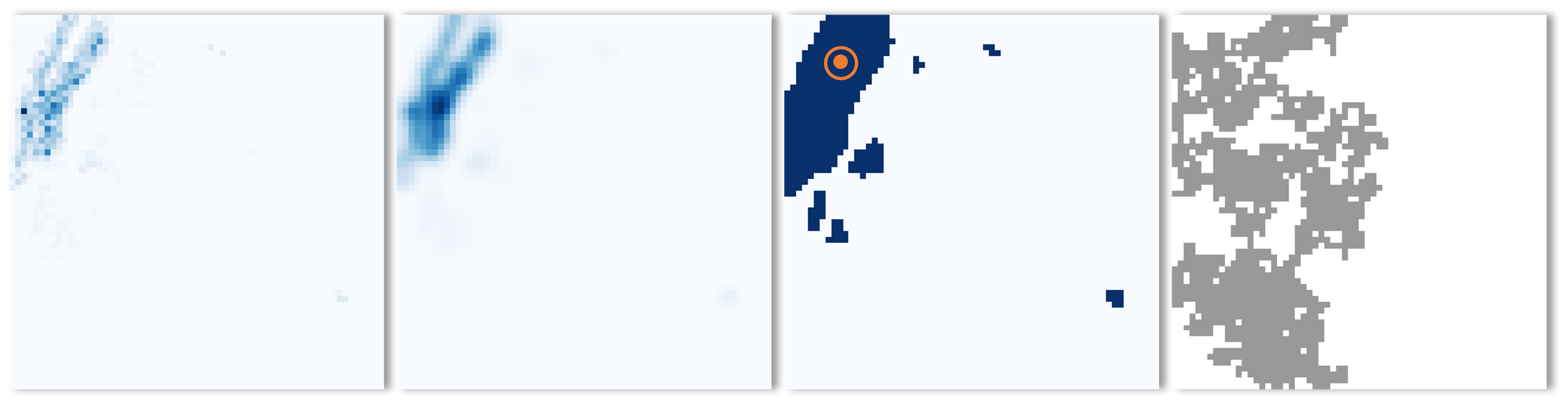}
         \caption{Biased masking. For each image (left), we first apply Gaussian blur and then threshold the image (middle images), then select a starting point at random in the thresholded region and grow a mask via random walk (right).}
         \label{biased_mask}
     \end{subfigure}
    \label{fig:masking}
    \caption{Comparison of the random and biased masking regimes.}
\end{figure}

By default, masks can be generated by randomly select a starting point in the image and then conducting a random walk for a fixed number of step.  We call this process \textbf{random masking}. However, since urban data is constrained by the built environment and is therefore highly skewed toward populated areas, random masks tend to include a large number of zero-valued cells, squandering opportunities to learn from the steep gradients in dense, high-traffic regions; Figure \ref{random_mask} illustrates  an example. To focus attention on populated areas, we use a \textbf{biased masking} approach: 1) Given an input image, apply Gaussian blur to blend the pixel values and increase the region of potential starting points. 2) Select a threshold (e.g., 90\% percentile of the image values) to identify populous regions. 3) Randomly select a starting location from one of the detected areas and generate masks via random walk. The probability of selecting one of the detected areas is proportional to the size of the area. These steps are illustrated in Figure \ref{biased_mask}. The biased masking approach makes the learning problem more challenging by increasing ``contrast'': ensuring that masks tend to include dense, dynamic regions, but also include sparse, stable regions. To compare the performance of the two masking approaches, we generated two masks (one random and one biased) for each training sample.

\section{Experimental Evaluation}\label{model}
\begin{figure*}[!btp]
    \centering
     \begin{subfigure}[b]{0.46\textwidth}
         \centering
         \includegraphics[width=\textwidth]{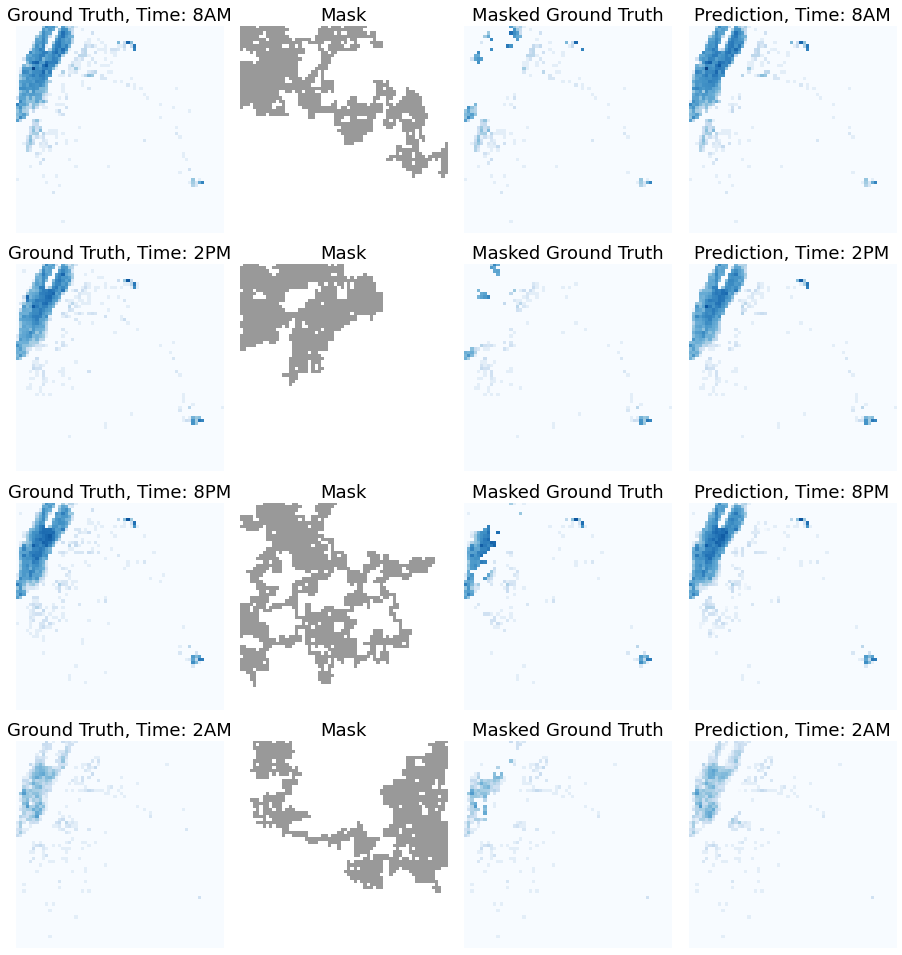}
         \label{fig:taxi_inpainting_example}
     \end{subfigure}
     \hspace{0.1cm}
     \begin{subfigure}[b]{0.46\textwidth}
         \centering
         \includegraphics[width=\textwidth]{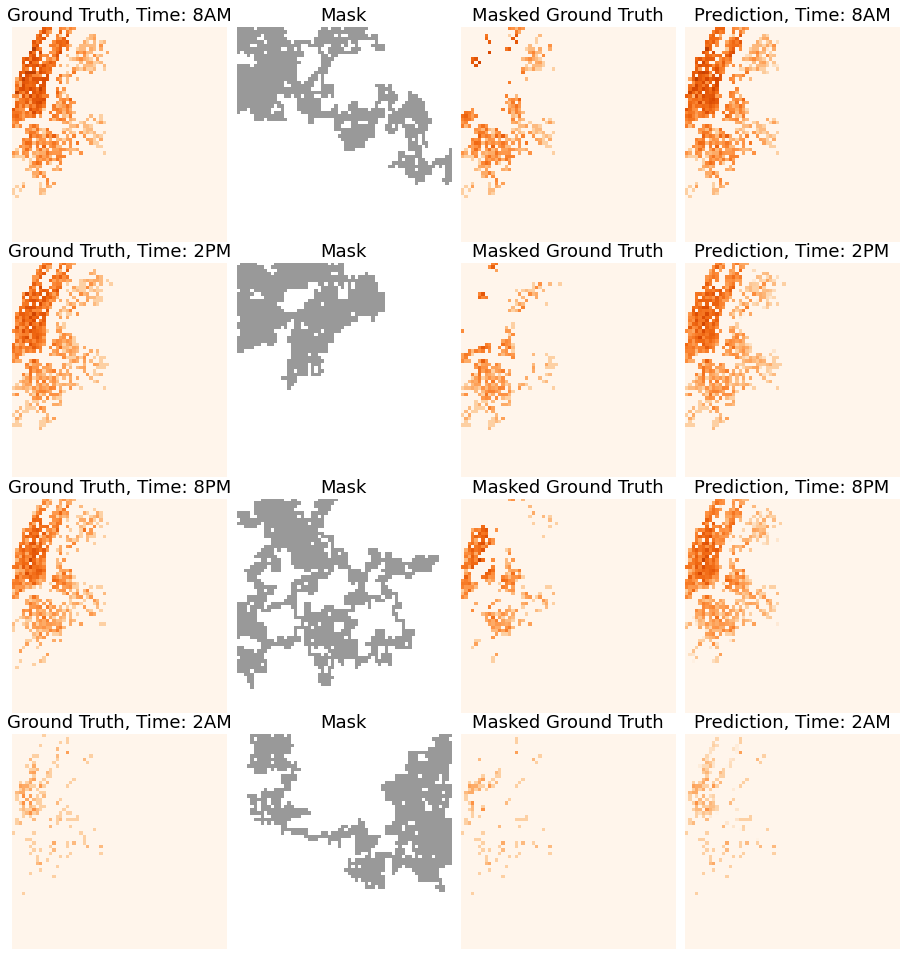}
         \label{fig:bikeshare_inpainting_example}
    \end{subfigure}
    \vspace{-0.7cm}
    \caption{Reconstructed results of taxi demand images (Left) and bike demand images (Right) at different hours time trained with biased masking and 3D partial convolutions (T=5 for taxi data and T=3 for bikeshare). From left to right, each column displays the ground truth image, mask, masked ground truth, and reconstructed data. From top to bottom, each row presents the taxi demand at 8AM, 2PM, 8PM, and 2AM, respectively.} 
    \label{fig:qualitative}
\end{figure*}

We consider the following questions:
\begin{enumerate}[leftmargin=0.27in,label=(Q\arabic*)]
    \item Is the core 3D model qualitatively \& quantitatively effective at inpainting missing data? (Section \ref{sec:qualitative}, Figure \ref{fig:qualitative}, Table \ref{tab:quan_results})
    
    \item Does increasing the number of timesteps per training sample generally improve performance? (Section \ref{sec:timesteps}, Figure \ref{fig:quan_results})
    
    \item Does biased masking improve performance overall, and in specific regions? (Section \ref{sec:effective_biased_masking}, Figure \ref{fig:convergence})
    
    \item Does varying the number of timesteps per training sample influence the spatial distribution of error between sparse and dense regions? (Section \ref{sec:timesteps}, Figure \ref{fig:sptial_error_dist})
    
    \item Does the model faithfully reconstruct local, dynamic conditions in specific areas of interest? (Section \ref{sec:scenario}, Figure \ref{fig:all_scenarios})
    
\end{enumerate}
With NYC taxi data, we trained the models on both mask types --- random and biased, and with different temporal dimension T = \{1,2,3,5,7,10,15\}. Based on initial experiments on both mask types and at lower temporal chunk sizes, we found that $\lambda=12$ offered effective performance; we fix $\lambda$ to be 12 for all experiments on the taxi data. The batch size and initial learning rate are set to 16 and 0.01 respectively. Learning rate decays every 500 training iterations at rate of 0.9. Unless otherwise stated, we evaluate the model on the test set using $\ell_{1,hole}$, which is the sum of the absolute value of the difference between the ground truth and predictions at the masked positions only. 

We compare our models with baseline statistical methods:
\begin{itemize}[leftmargin=0.12in]
    \item \textbf{Temporal Global Mean}: On the training data, we calculate the average taxi demand at each pixel, for each hour of the day. On the test data, we assign each masked pixel the corresponding global mean computed from the training data.
    \item \textbf{Nearest Neighbor (NN) Interpolation}: We assign each masked pixel the value of the nearest unmasked pixel. We experimented with both 2D and 3D implementations using scipy.\footnote{\url{https://docs.scipy.org/doc/scipy/reference/generated/scipy.interpolate.griddata.html\#scipy.interpolate.griddata}}
    \item \textbf{RBF Interpolation} We interpolate using radial basis functions (RBF) on observations at points sampled outside the masked region. We experimented with both 2D and 3D RBF interpolation with RBF Python implementation.\footnote{\url{https://github.com/treverhines/RBF}}
\end{itemize}
We considered 3D kriging, but found the poor scalability to be prohibitive: the estimated time to complete the computation for an experiment with T=2 was about two weeks on a typical platform.  Moreover, kriging is a linear method, and we have no reason to believe that it can reconstruct data across large, irregular regions.

Another approach, which we did not study, is to use physics-based models based on computational fluid dynamics~\cite{BLOCKEN2015219} or agent-based models that directly encode human behavior \cite{individual, SOUZA2019858} to capture macro traffic dynamics.  These approaches can potentially "fill" large missing regions, but must be designed separately for each application rather than learned automatically from data.



\subsection{Model Effectiveness (Q1)}
\label{sec:qualitative}

We find that for both taxi and bikeshare datasets the proposed model faithfully captures qualitative visual patterns and also significantly outperforms baseline methods on multiple metrics.

\subsubsection{Qualitative Analysis} We first present some visual examples of inpainting results on NYC taxi data in Figure \ref{fig:qualitative}. The left figure shows taxi demand at four different hours of the day (8AM, 2PM, 8PM, and 2AM). From left to right, we show the ground truth, the (biased) mask, the mask applied to the ground truth, and the reconstructed image. The inpainting model was trained with 5 timesteps per training sample and with biased masking. 

For all hours and all masks, the model is effective at reconstructing missing data, even when the majority of the signal is obscured.  The reason is clear: the patterns are sufficiently stable from timestep to timestep as to allow the model to infer missing values from temporal patterns as well as spatial patterns.  The model is also responsive to the time of day: We see fewer rides at 2AM than at 2PM, as expected, suggesting that the model has learned temporally local patterns as opposed to relying on global spatial patterns. The transition across the mask boundary is also smooth, suggesting the model was able to consider local spatial patterns appropriately.  Overall, we find that the model is perceptually effective at reconstructing missing values, even in challenging cases.

The right plot in Figure \ref{fig:qualitative} visually shows corresponding results for bikeshare data. The model was trained with bikeshare data using T=3, biased masking and $\lambda = 4$. We observe similar observations as the results from taxi data --- at all times of day and for all masks,  the reconstructed images are visually similar to the ground truth images, indicating the consistent effectiveness of our model.  

\subsubsection{Quantitative Analysis} Table \ref{tab:quan_results} contains quantitative results of baseline models and our neural models in different evaluation metrics. We observe that: 1) Our neural models, trained with either masking type or with any temporal dimension, always outperform the baseline models. The 2D baseline models that ignore the temporal dimension are especially ineffective. Global mean ignores spatial effects and just models a function $pixel, hour \rightarrow value$.  2D- and 3D- nearest neighbor methods perform poorly when the nearest neighbors may be far away; 2D- and 3D-RBF methods assume relatively uniform sampling across the region, which is not possible in our setting of wide-area missing data. 2) At T=5 and 7, our method performs similarly and achieves the best performances --- almost 50\% lower $\ell_1$ error and 66\% lower $\ell_2$ error than the best baseline. 3) SSIM does not significantly distinguish different models; while popular in image inpainting, this metric is designed to capture perceptual similarity of natural images, which are not relevant for the spatiotemporal aggregations we study. 4) The model training time increases by about 9 minutes for every additional hour included in a chunk. At T=5, the model takes 55 minutes to train. The baseline heuristic-based methods --- global mean and 2D- and 3D-NN --- are very fast (completing in a few minutes) but very inaccurate given that they do not attempt to model global dynamics. The 3D-RBF method is inefficient: T=2 required over 24 hours to train.
\begin{table}[!hb]
    \footnotesize
    \tabcolsep=0.13cm
    \renewcommand*{\arraystretch}{1}
    \centering
    \begin{tabular}{p{1.3cm}cccccc}
        
        \textbf{Model} & \textbf{Mask Type} & \textbf{$\ell_{1,hole}$} & \textbf{$\ell_{2,hole}$} & \textbf{SSIM} & \textbf{PSNR} & \textbf{Train (m)}\\ \hline
        Global Mean & - & 1.2644 & 55.3298 & 0.9973 & 61.4880 & <5 \\ 
        2D-RBF & - & 3.1442 & 284.8807 & 0.9890 & 54.8346 & 70 \\ 
        2D-NN & - & 3.1179 & 318.6575 & 0.9884 & 54.0717 & <5 \\
        3D-RBF & - & 1.6653 & 94.7708 & 0.9956 & 57.9921 & >24h \\
        3D-NN & - & 1.3632 & 84.0529 & 0.9964 & 59.1652 & <5 \\ \hline
        
        \multirow{2}{*}{Ours, $T=1$} & biased & 0.9081 & 37.8468 & 0.9984 & 62.4268 & 18 \\
        & random & 0.9406 & 40.3730 & 0.9983 & 62.4679 & 18\\ \hline
        
        \multirow{2}{*}{Ours, $T=2$} & biased & 0.8551 & 32.6429 & 0.9986 & 63.1815 & 27\\
        & random & 0.8979 & 35.2923 & 0.9985 & 63.1056 & 27\\ \hline
        
        \multirow{2}{*}{Ours, $T=3$} & biased & 0.7847 & 25.8374 & 0.9987 & 63.6445 & 35 \\
        & random & 0.7950 & 26.4765 & 0.9989 & 63.7221 & 35\\ \hline
        
        \multirow{2}{*}{Ours, $T=5$} & biased & 0.7196 & 18.7080 & \textbf{0.9991} & \textbf{64.4028} & 55 \\
        & random & 0.7606 & 20.6116 & 0.9990 & 64.1000 & 55\\ \hline
        
        \multirow{2}{*}{Ours, $T=7$} & biased & \textbf{0.7185} & \textbf{18.6746} & 0.9990 & 64.3407 & 75\\
        & random & 0.7489 & 20.0100 & 0.9990 & 64.2656 & 75\\ \hline
        
        \multirow{2}{*}{Ours, $T=10$} & biased & 0.7537 & 24.8383 & 0.9986 & 63.3329 & 75\\
        & random & 0.7820 & 26.1138 & 0.9985 & 63.1288 & 75\\ \hline
        
        \multirow{2}{*}{Ours, $T=15$} & biased & 0.7729 & 25.3386 & 0.9985 & 63.1885 & 140\\
        & random & 0.7849 & 21.9446 & 0.9989 & 63.8721 & 140\\ \hline
    \end{tabular}
    \caption{Model training time and performance.}
    \label{tab:quan_results}
    \vspace{-2em}
\end{table}

\subsection{Temporal Dimension Tradeoff (Q2)}\label{sec:timesteps}
\begin{figure}[!ht]
    \centering
        \includegraphics[width=0.46\textwidth]{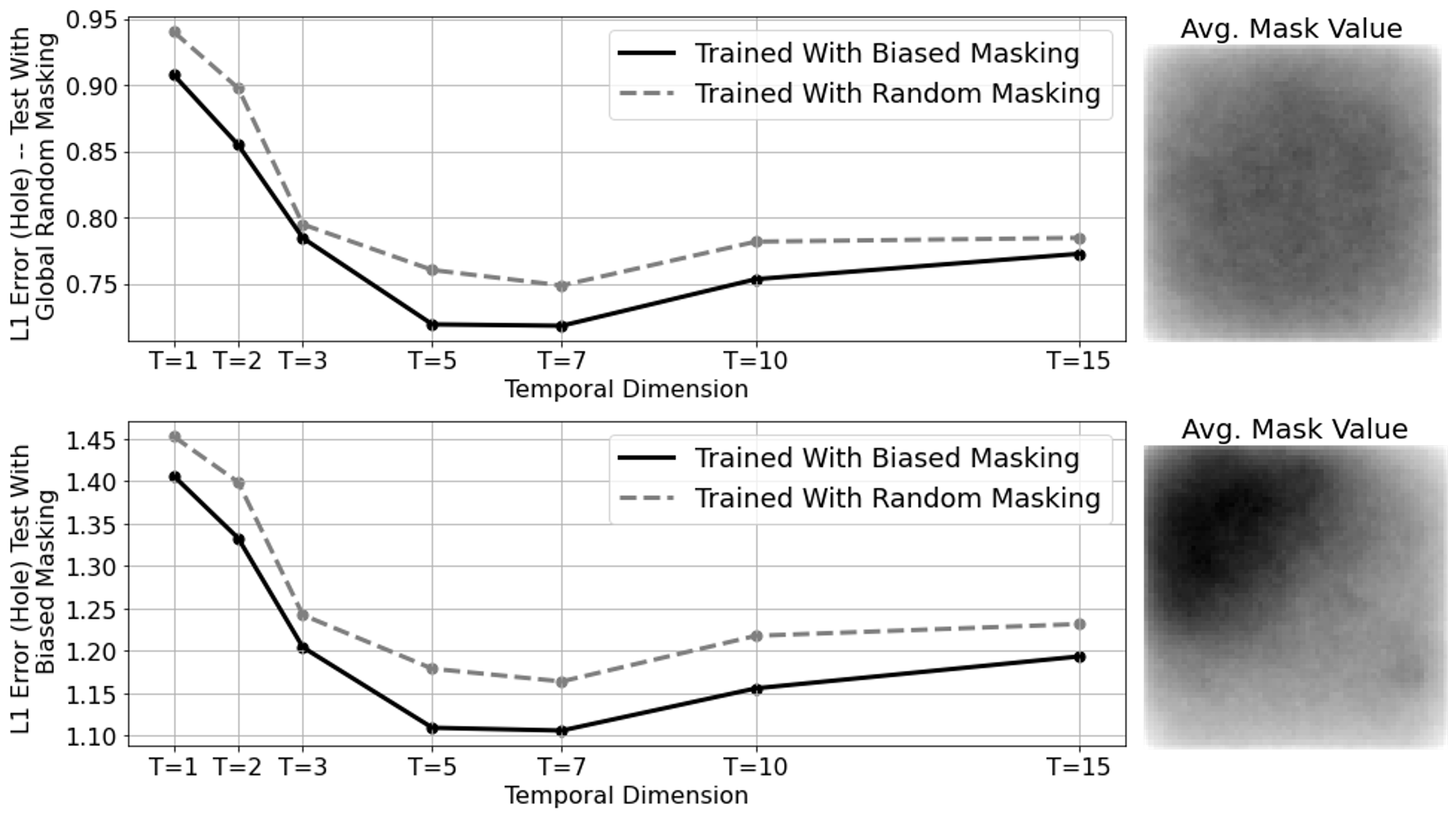}
        \vspace{-0.2cm}
        \caption{Evaluation of models trained with biased masking against those trained with random masking, at seven temporal dimensions, with two different masking scenarios --- random and biased masking.}
    \label{fig:quan_results}
\end{figure}

Figure \ref{fig:quan_results} shows the prediction errors for NYC taxi data, evaluated on random masks (top plot) and biased masks (bottom plot). The y-axis is the $\ell_1$ loss considered for the masked region only ("Hole"). The x-axis varies the number of timesteps included per training sample (Temporal dimension), ranging from 1 to 15. (a) When tested with random masks, the average mask covers the entire region, concentrated at the center. Models trained with biased masking reduces error at all sizes. The $\ell_1$ error decreases as the number of timesteps increases up until T=7, then starts to increase again (T=5 and T=7 have similar performances when trained with biased masking.) At T=2, the model begins to make use of the temporal dependency between the data by applying 3D convolutions. With both biased and random masking, the $\ell_1$ loss decreases sharply when T changes from 1 to 5. (b) When tested with biased masks, the average masked cells are concentrated at the upper left due to the bias toward populated regions. The plot has a similar U-shape as that of random masking.

\subsection{Biased Masking is Effective (Q3)}\label{sec:effective_biased_masking}
Figure \ref{fig:quan_results}, as discussed, compares the effects of biased masking to random masking at various value of T; we see that at all tested temporal dimensions, models trained with biased masking outperform those trained with random masking, indicated by smaller $\ell_1$ errors.

\begin{figure}[!ht]
    \centering
        \includegraphics[width=0.46\textwidth]{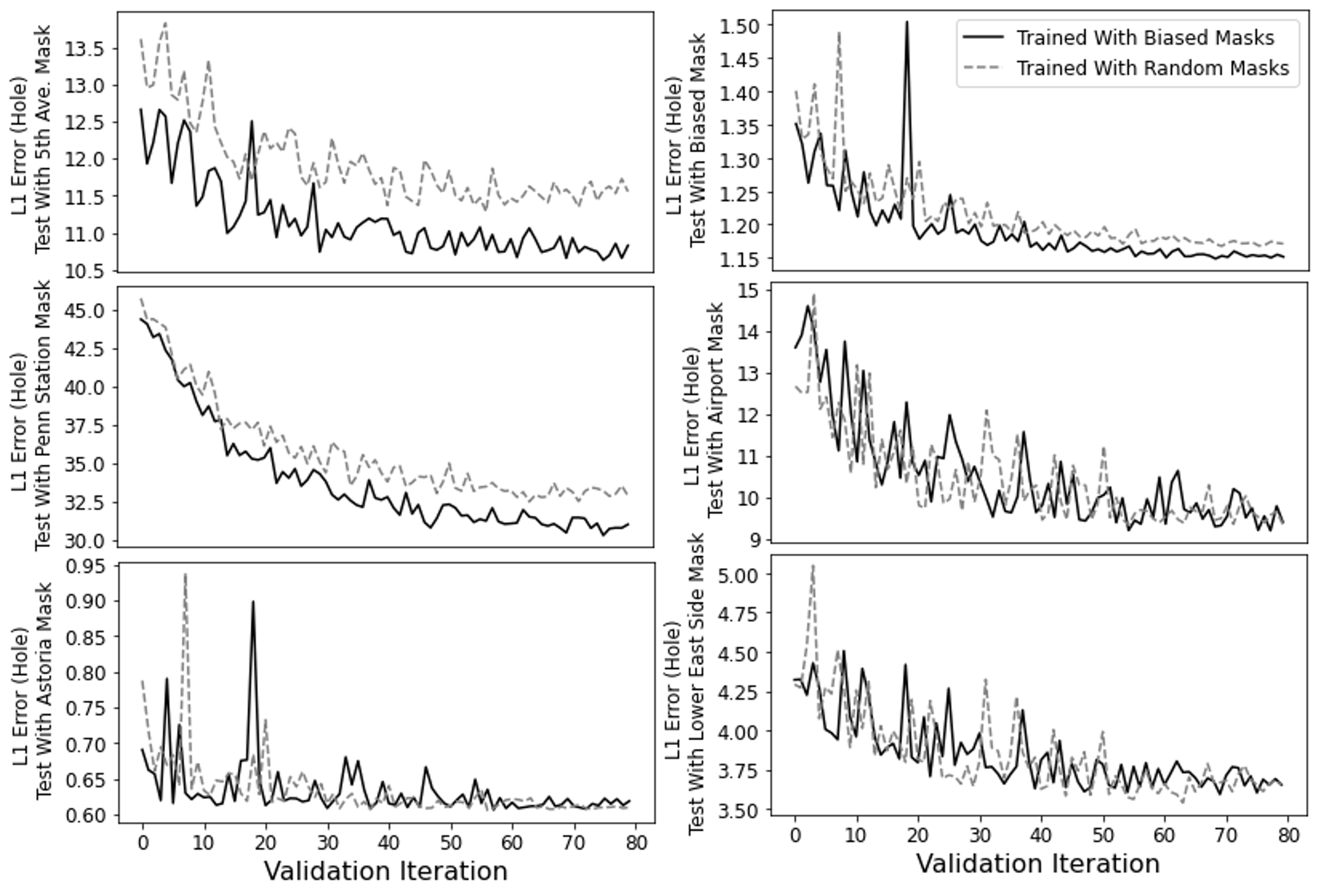}
        \caption{Convergence plots of the models trained with either biased or random masking, and tested with random masks, biased masks and other five additional scenarios maskings.}
    \label{fig:convergence}
\end{figure}

In addition to the measurement of overall error, we also inspected the convergence rates under both training regimes, as measured by the validation set with our selected scenarios (Figure \ref{fig:convergence}). The scenario masks are chosen to evaluate local accuracy in high-traffic, low-traffic, high-variability, and semantically important locations. See \ref{sec:scenario} for masks of the scenarios and detailed evaluations.

Overall, when we tested with random and biased masks, the model trained with biased masks converged faster and had smaller errors, indicating that biased masking is beneficial to the imputation task under skewed distributions (upper left). Evaluating the 5th Avenue and Penn station scenarios, the model trained with biased masking displayed similar patterns --- they converged faster and achieved better results than the model trained with random masks. Those two scenarios are representative of dense and busy areas.  We conjecture that biased masking avoids rewarding the model for trivially predicting zero in sparse regions and ignoring the dynamics in dense regions.   We consider this result an initial foray: encoding domain knowledge and data patterns into the masking strategy appears to be a powerful, easy, and architecture-agnostic means of improving model performance, aligned with emerging principles of data-centric AI. The other three scenarios --- airport, lower east side, and Astoria, represent sparse regions with relatively light traffic. The convergence lines for them are less stable, and no benefit of biased masking is realized.  We conjecture that variants of biased masking to weight both dense and sparse (yet non-zero) areas may further improve the model, as would specialized training on regions of interest (though that approach could be considered data leakage from training to test).
\vspace{-0.2cm}
\vspace{-0.2cm}
\begin{figure}[!ht]
    \centering
        \includegraphics[width=0.46\textwidth]{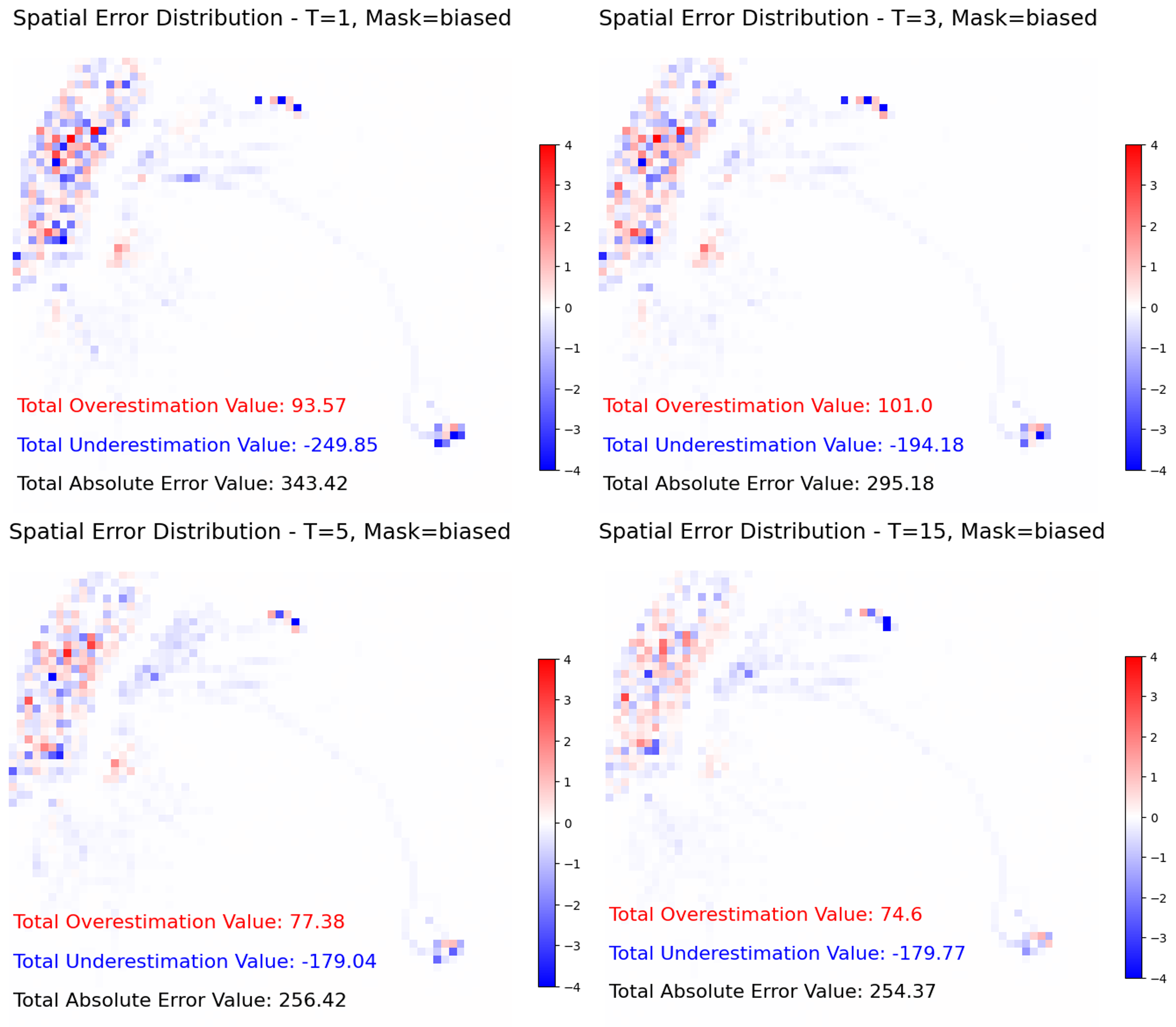}
        \caption{Aggregated spatial errors between predicted and ground truth values, from models trained with different temporal dimensions. Red areas indicate overestimation, while blue areas represent underestimation.}
    \label{fig:sptial_error_dist}
    \vspace{-1em}
\end{figure}

\subsection{Spatial distribution of errors (Q4)} 
We hypothesized that the original 2D partial convolution architecture (corresponding to T=1, Figure \ref{fig:quan_results}(a)) would be insufficient to capture transient events. For example, taxi rides occur in the suburbs, but they are infrequent and less predictable; we expected the model to be less capable of accurately predicting these events. Increasing the temporal dimension is also expected to be helpful with the dense region as well.

We can inspect the spatial distribution of the error for T=1 in Figure \ref{fig:sptial_error_dist} to check this hypothesis: Each map is the average of 3000 timesteps, and is colored by the difference between the predicted value and the ground truth: a blue cell indicates an underestimate and a red cell represents an overestimate.  We see that the suburban regions are consistently underestimated, while the dense region is overestimated. At T=5, we observe similar pattern, but with both underestimation and overestimation errors significantly reduced. The suburbs are still underestimated, but the dense regions are effectively improved when more temporal dimensions are incorporated. At T=15, the spatial error distribution is almost identical to T=5, with slightly higher underestimation and lower overestimation. However, T=15 requires prohibitive training time due to very large training samples, so this approach is undesirable with just slightly better performance. This tradeoff in temporal scope reflects a subtle characteristic of the source data; we hypothesize that T=5 corresponds to the window size needed to capture dynamic traffic periods; e.g., morning and evening commutes. 

\subsection{Scenario Based Evaluation (Q5)}\label{sec:scenario}
\begin{figure}[hb]
    \centering
        \includegraphics[width=0.48\textwidth]{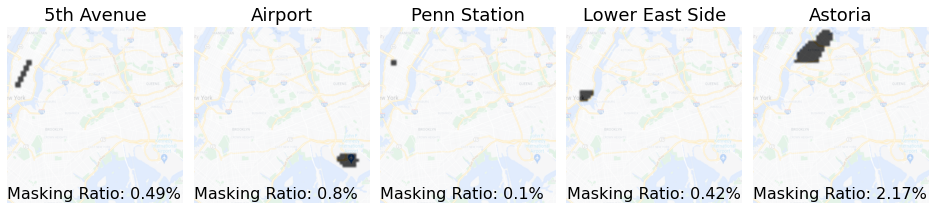}
        \caption{Scenario masks overlaid on NYC map. Annotation: The ratio of masked-to-unmasked area.}
    \label{fig:scenario_masks}
\end{figure}

Spatiotemporal patterns of missing data in practice are unlikely to resemble random walks. Instead, outages will correlate with environmental features: sensors may fail in certain weather conditions, transient events may prevent data acquisition, or legal restrictions on data availability may follow political boundaries. To demonstrate the applicability of our inpainting models in real-world situations, we evaluate the inpainting methods based on specific locations representing varying conditions. We tested five different scenarios to cover various spatial locations, temporal variances, and social events. The five scenarios include the masking of 5th Avenue, Penn Station, airport, lower east side, and Astoria. The masks are visualized in Figure
\ref{fig:scenario_masks}.
\vspace{-0.3cm}

As mentioned in Section \ref{sec:effective_biased_masking}, 5th Avenue and Penn station are representative of busy and dense areas with heavy traffic. 5th Avenue can also show the impacts of certain social events on traffic patterns: The Pride Parade showed an anomalous intervention where traffic was zero on the parade route. Lower East Side is away from central Manhattan, with relatively lighter traffic than the first two cases. The scenario of airport and Astoria represent the sparse regions where traffic is light.

We chose two periods for those scenarios to cover temporal variance -- Feb. 1st to Feb. 15th, 2016, and June, 18th to June 29th, 2016. A snowstorm from Feb 5th to 8th in New York City is evident in the data (Figure~\ref{fig:all_scenarios}). 
On June 26th, 2016, the Pride Parade in New York City started at 5th Avenue, and moved downtown to 8th Street. The event blocked all traffic along the route and affected the surrounding traffic as well. Therefore, testing in the selected June period can help evaluate the model's response to anomalies.

We test three inpainting models --- our model trained with biased masking at T=5, the same model but trained with random masking at T=5, and the global mean approach. We plotted the ground truth and predicted values at the average pixel level in the missing region, for each hour during the selected periods. The visualizations are provided in Figure \ref{fig:all_scenarios}. The average absolute errors between the ground truth and predicted values, over the missing region and during the evaluation periods, are reported in Table \ref{tab:scenario-errors}. We have the following observations:
\begin{figure}[ht]
    \centering
        \includegraphics[width=0.46\textwidth]{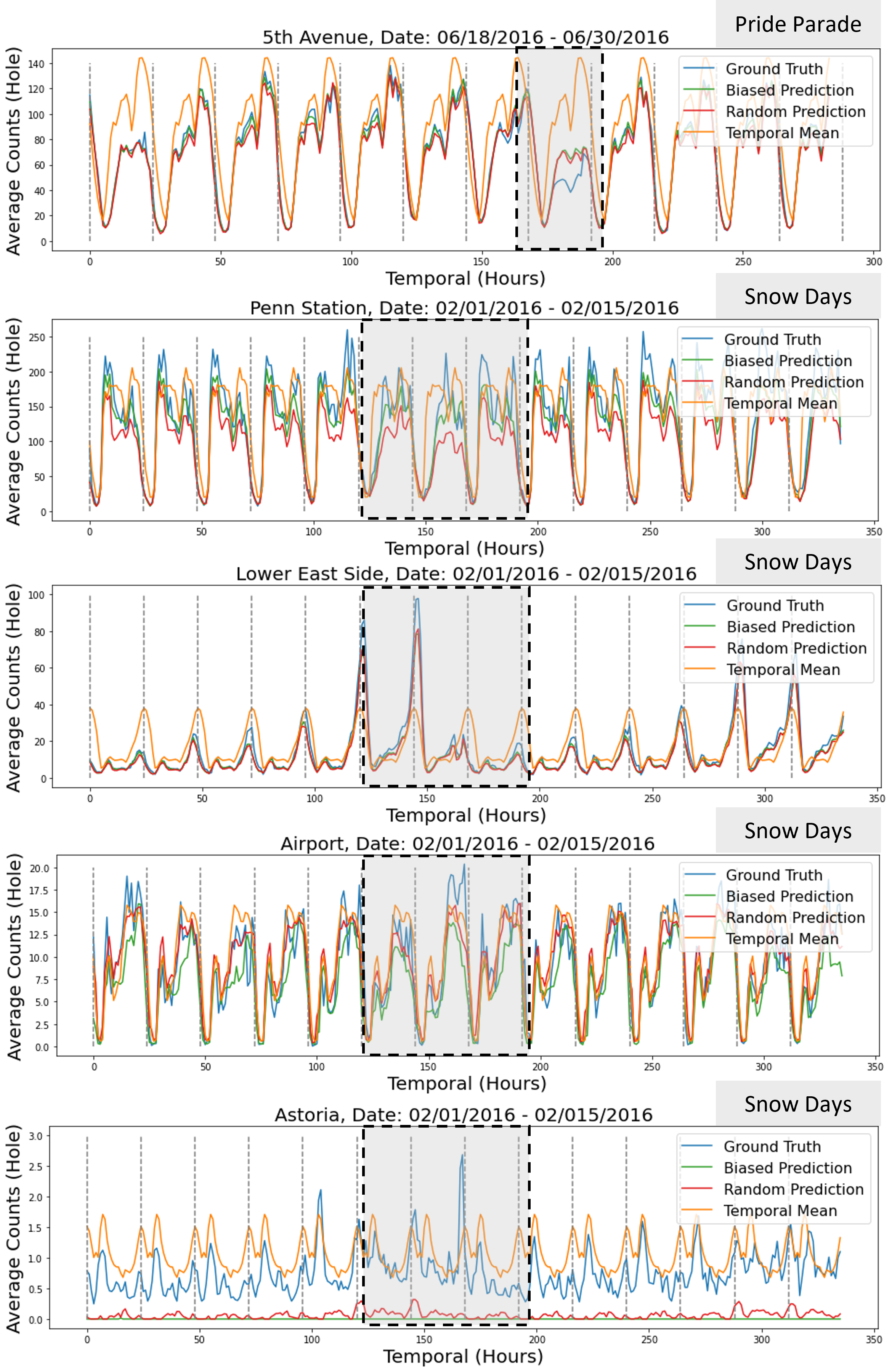}
        \caption{Temporal line plots of evaluations for five scenarios. In each plot, we visualize the ground truth, prediction from model trained with biased masking and random masking, and predictions from temporal mean method. Two evaluation periods, Feb. and June are selected. The irregular events, extreme snow days and pride parade, are annotated with grey regions.}
    \label{fig:all_scenarios}
\end{figure}
\begin{table}[ht]
    \footnotesize
    \tabcolsep=0.2cm
    \renewcommand*{\arraystretch}{1}
    \centering
    \begin{tabular}{cccc}
        \textbf{Scenarios} & \textbf{G.T.- Biased} & \textbf{G.T. - Random} & \textbf{G.T. - Mean} \\ \hline\hline
        \multicolumn{4}{c}{\textbf{02/01/2016 --- 02/15/2016}} \\ \hline
        5th Avenue & 4.2 & 6.2 & 17.0 \\ \hline
        Penn Station & 19.3 & 33.5 &  30.0\\ \hline
        Lower East Side & 2.5 & 2.8 & 8.2\\ \hline
        Airport & 2.3 & 1.6 &  1.8\\ \hline
        Astoria & 0.8 & 0.7 & 0.4\\ \hline
        \multicolumn{4}{c}{\textbf{06/18/2016 --- 06/30/2016}} \\ \hline
        5th Avenue & 3.6 & 4.8 & 22.53\\ \hline
        Penn Station & 21.6 & 37.5 &  30.0\\ \hline
        Lower East Side & 1.7 & 2.1 & 7.4\\ \hline
        Airport & 2.4 & 1.9 &  2.0\\ \hline
        Astoria & 0.8 & 0.7 & 0.4\\ \hline
    \end{tabular}
    \caption{Average absolute error between the predicted values and ground truth, over the missing regions, and during the selected evaluation periods.}
    \label{tab:scenario-errors}
\end{table}
\vspace{-0.3cm}
\begin{itemize}[leftmargin=0.12in]
    \item For three scenarios --- 5th Avenue, Penn Station, and Lower East Side, our models --- whether trained with biased or random masking --- have much smaller gaps between the predicted values and the ground truth, compared with the temporal mean approach. This benefit holds for both evaluated periods, as shown in both Table \ref{tab:scenario-errors} and Figure \ref{fig:all_scenarios}. For the airport and Astoria scenarios, the temporal mean is slightly better, with much smaller magnitude in comparison with other three cases.
    
    \item rom Table \ref{tab:scenario-errors}, we see that for both evaluation periods, the model trained with biased masking has smaller average errors than the model trained with random masking, other than the scenario of airport during June. 
    \item  During the snow days (02/05-02/08/2016), it is expected that the traffic in the dense regions would be significantly impacted, which can be supported by the trough seen from the ground truth line in the scenario of Penn Station (other scenarios are not heavily impacted by the snow.) The model trained with biased masking is responsive to the irregular traffic caused by extreme weather, unlike the temporal mean baseline.
    
    \item During the event pride parade, the traffic on 5th Avenue was all diverted to other routes, creating an anomaly in the traffic patterns. Therefore, we saw a dip in the traffic counts. Similar observation as the snow day, the temporal mean baseline does not recover the missing values . However, even though the inpainting results from our model are close to the ground truth values, they slightly overestimate the results.
\end{itemize}

\begin{figure}[ht]
    \centering
        \includegraphics[width=0.45\textwidth]{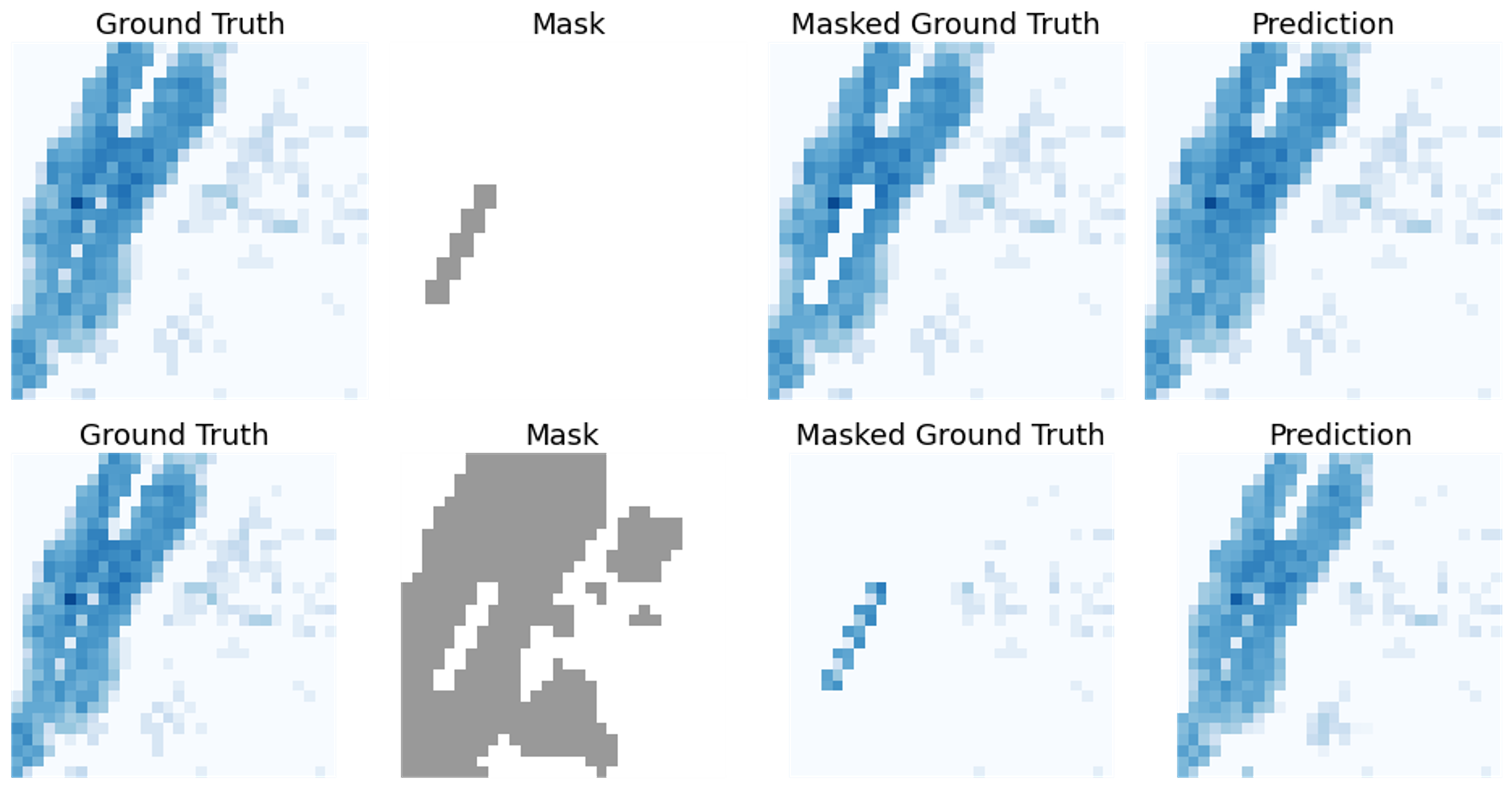}
        \caption{Top: ``Airbrushing'' the parade event (white pixels) to remove its effect on the data. Bottom: Inferring traffic effects of the parade by reconstructing data everywhere except 5th Avenue to produce qualitatively realistic results.}
    \label{fig:parade}
\end{figure}

Overall, the reconstruction accuracy is compelling at specific locations, but not perfect. For 5th Avenue scenario, the parade can be seen as an anomaly, which is rare in the training stage and hard to be detected. But  this scenario represents another application usage of our model: rather than assuming that ground truth data is ``correct". We use the masking to intentionally repair known bad data, and reconstruct global patterns in a semantically reasonable way. This ``airbrushing'' of flaws in the data can be used to improve the quality of training sets for downstream applications, such as biofouled or errant sensors and faulty telemetry. For example, from the top visualization in Figure \ref{fig:parade}, we visualize the 5th Avenue scenario: The first column shows the taxi counts along 5th Avenue during parade day, zoomed in on the Manhattan region. Several locations of missing data (white dots) can be seen on the avenue. We masked out the 5th Avenue altogether and used our inpainting model to reconstruct the missing values. The use case is to enable policymakers and researchers to conduct counterfactual studies: what would have taxi demand been like were it not for the parade?  The results, as shown in the forth column, recover the missing regions in a realistic way.

Alternatively, the model might be used to synthesize parade-day traffic rather than removing its effects.  By masking the surrounding area and retaining the parade disruption, the model can attempt to represent the influence of the disruption elsewhere in the city.  As shown from the bottom visualization in Figure \ref{fig:parade}, the generated results are smaller in magnitude, but overall the pattern is matched faithfully, suggesting this use case is viable for synthesizing scenarios that may not be present in the data record (natural disasters, proposed construction, accidents, etc.).
\vspace{-0.2cm}
Penn Station is a train station and represents a high-demand area for taxis. Our model tends to underestimate the high demand at this location, though biased masking improves the prediction. For Lower East Side, there are a few anomalous spikes, to which the proposed models are responsive. For airport and Astoria, our models are no better than the temporal mean approach. We conjecture that for airport, the highly variable rides in and out of the airport confound the model. For Astoria, the much lower demand is harder to predict; note the lower scale of the y-axis.

\section{Discussion}

Our study is motivated by the inconsistent availability of urban data caused by missing, corrupt, or inaccurate data, which hinders their use in downstream tasks, especially learning tasks, that require coverage and accuracy. We designed and implemented a model based on partial convolutions that can tolerate irregular missing regions --- zip codes, geographical boundaries, congrssional districts, or other regions that may correlated with data absence or quality.  To capture the temporal dependency in urban data, we replaced 2D convolutional layers in the model with 3D convolutional layers and experimented with varying the number of timesteps per training sample, finding non-trivial tradeoffs and a local optimum around T=5 for taxis and T=3 for bikeshare, potentially interpretable as the autocorrelation period of traffic (i.e., about 5 hours of rush hour). 

To address the spatial skew in human activity, we proposed a masking approach that can reflect the skew in the distribution. By encouraging the model to attend to dense, dynamic regions (via a percentile threshold), the model learns faster and is not rewarded for accurate predictions in trivially inactive areas.  Biased masking showed improved performance across all values of $T$, multiple global evaluation strategies, and most local evaluation scenarios. This approach suggests a broader family of related masking strategies to help users encode domain knowledge about the data and setting.  For example, encoding correlations between high-traffic areas (e.g., subway stops and train stations during lunch time) as masks may help the model learn these correlations with less data.

Qualitatively, we confirmed from the visual examples that image inpainting techniques can be used to reconstruct data in large, irregular regions in space and time. Quantitatively, we confirmed that extending the model architecture to 3D benefits improves performance, as supported by the sharp decrease in $\ell_1$ when T changes from 1 to 2. Second, we observe that increasing the temporal dimension to a certain threshold improves performance in general, regardless of masking strategy; ignoring the temporal dimension in this setting is untenable. 

Additionally, we evaluated performance in local settings, demonstrating that the model is not just learning an average value, but is responsive to subtle spatial variation. The model captures irregular traffic patterns caused by transient events, such as extreme weather and the Pride Parade, and showed that biased masking can improve performance in local settings. Additionally, the scenario evaluations also showcased the better results introduced by the biased masking than the random masking.

\section{Limitations \& Future Work}
There are several limitations of our study that represent directions for future work. First, our results on mobility data may extend to other urban activity (e.g., 311 calls, crowd movement, business permits, public safety events, housing events, and more). We do not consider the generalizability of these methods to multiple variables, or variables that do not follow the same spatial patterns; there are opportunities to exploit correlations between variables to improve performance. Additionally, the taxi dataset is exceptionally large and complete; understanding how these techniques behave in low-data regimes is important for practical applications. Integration of masked multi-variate data may be an opportunity: given the shared built environment, models trained on one variable may transfer to predictions of other variables.  Second, rasterizing event data to a form amenable to computer vision techniques involves a number of design choices we did not study: resolution, overlap, and irregular boundaries may present opportunities or challenges.  In particular, data associated with census blocks, tracts, or individual trajectories lose information when regridded as histograms. In these cases, graph neural networks may be more appropriate to represent the spatial adjacency relationships. Third, even with the best model configuration, we consistently overestimate in the city region and underestimate in the sparse suburban region. Some model architectures (attention mechanism, multi-view learning) or loss functions may improve performance, as may more specialized masking and training regimes. 

\section{Code Availablity}
Our code is available at [anonymized for review].


\bibliographystyle{ACM-Reference-Format}
\bibliography{inpainting}

\end{document}